\documentclass[10pt,twocolumn,letterpaper]{article}

\usepackage{iccv}              
\usepackage{pifont}
\newcommand{\xmark}{\ding{55}}%
\newcommand{\cmark}{\ding{51}}%
\usepackage{multirow}
\definecolor{iccvblue}{rgb}{0.21,0.49,0.74}
\usepackage[pagebackref,breaklinks,colorlinks,allcolors=iccvblue]{hyperref}

\def\paperID{*****} 
\def\confName{ICCV}
\def\confYear{2025}

\title{Learning Point Cloud Representations with Pose Continuity for Depth-Based Category-Level 6D Object Pose Estimation} 

\author{ \\ 
Zhujun Li$^{1}$\ \ \ \ \ \ Shuo Zhang$^{2,3}$\ \ \ \ \ \ Ioannis Stamos$^{1,2}$\\
$^1$Graduate Center, CUNY\ \ \ \ \ \ $^2$Hunter College, CUNY\ \ \ \ \ \ $^3$Weill Cornell Medicine\\
{\tt\small zli3@gradcenter.cuny.edu\ \ \ \ \ \ \{sz780, istamos\}@hunter.cuny.edu}
}
\begin{document}
\maketitle
\begin{abstract}
Category-level object pose estimation aims to predict the 6D pose and 3D size of objects within given categories. 
Existing approaches for this task rely solely on 6D poses as supervisory signals without explicitly capturing the intrinsic continuity of poses, leading to inconsistencies in predictions and reduced generalization to unseen poses. To address this limitation, we propose HRC-Pose, a novel depth-only framework for category-level object pose estimation, which leverages contrastive learning to learn point cloud representations that preserve the continuity of 6D poses. HRC-Pose decouples object pose into rotation and translation components, which are separately encoded and leveraged throughout the network. Specifically, we introduce a contrastive learning strategy for multi-task, multi-category scenarios based on our 6D pose-aware hierarchical ranking scheme, which contrasts point clouds from multiple categories by considering rotational and translational differences as well as categorical information.
We further design pose estimation modules that separately process the learned rotation-aware and translation-aware embeddings. Our experiments demonstrate that HRC-Pose successfully learns continuous feature spaces. Results on REAL275 and CAMERA25 benchmarks show that our method consistently outperforms existing depth-only state-of-the-art methods and runs in real-time, demonstrating its effectiveness and potential for real-world applications. Our code is at \url{https://github.com/zhujunli1993/HRC-Pose}.
\end{abstract}
    
\section{INTRODUCTION}
\label{sec:intro}

\begin{figure}[h]
\begin{center}
\includegraphics[width=0.47\textwidth]{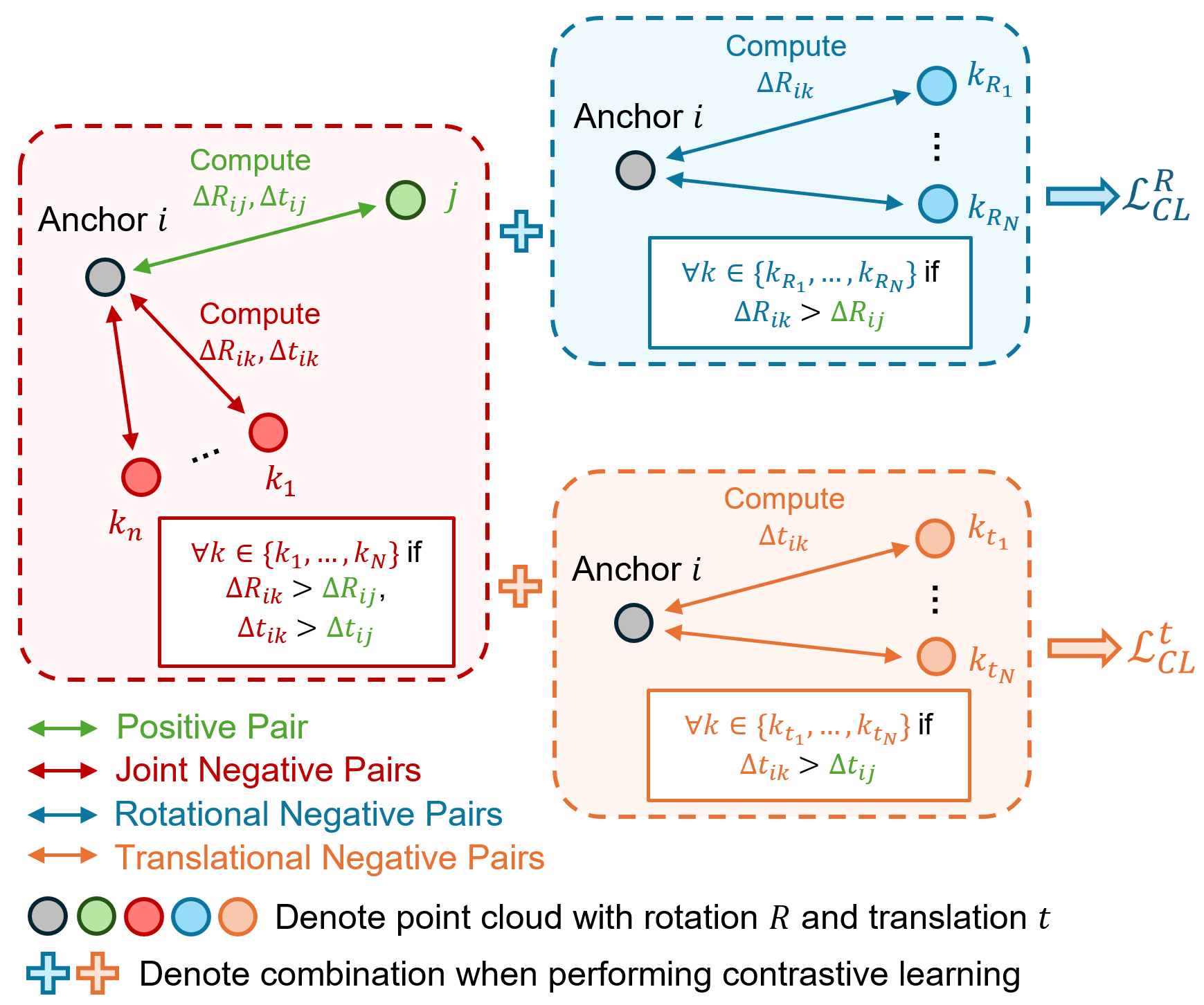}
\caption{{Illustration of our contrastive learning strategy with 6D pose-aware hierarchical ranking. } $\Delta$ denotes pose distance. Within each category, for an anchor point cloud indexed by $i$, we form positive pairs by randomly selecting another point cloud $j$ from the batch. Given a positive pair $(i,j)$ with pose distances $\Delta R_{ij}$ and $\Delta t_{ij}$, a sample indexed by $k$ in the batch forms a joint negative pair with $i$ if $\Delta R_{ij}<\Delta R_{ik}$ and $\Delta t_{ij}<\Delta t_{ik}$, as shown in the red box. A larger $\Delta R_{ik}$ and $\Delta t_{ik}$ indicate a higher ranking for $k$ in the batch. Additionally, samples are ranked separately based on $\Delta R$ or $\Delta t$, as shown in the blue and orange boxes. The final losses $\mathcal{L}^{R}_{CL}$ and $\mathcal{L}^{t}_{CL}$ are computed based on the combination of joint and rotational/translational-specific pairs.}
\label{p_1}
\end{center}
\vspace{-0.8cm}
\end{figure}

Category-level pose estimation aims to estimate the 6D poses and sizes of arbitrary objects within predefined categories without requiring exact CAD models. In contrast, instance-level pose estimation \cite{peng2019pvnet, yu20206dof, gao20206d, he2021ffb6d, cai2022ove6d, li2023depth} relies on CAD models of target objects, limiting its applicability to known instances. By removing this dependency, category-level pose estimation enables broader generalization across object instances, making it more adaptable to real-world scenarios such as robotic manipulation \cite{deng2020self}, scene understanding \cite{chen2019holistic++}, autonomous driving \cite{wu20196d}, and augmented reality \cite{su2019deep}.

Earlier methods \cite{wang2019normalized, Tian_2020_ECCV, chen2021sgpa, wang2021category, zhang2022rbp, zhang2022ssp, liu2023net, wang2023query6dof} tackle category-level pose estimation by predicting a normalized object coordinate space (NOCS) map and establishing pixel-wise correspondences. However, these approaches are highly sensitive to errors in 2D-3D correspondences or noisy depth inputs. To address these limitations, recent methods \cite{chen2020category, chen2021fs, lin2021dualposenet, lin2022category, di2022gpv, li2023generative, zheng2023hs} bypass correspondence construction by directly estimating poses, simplifying the pipeline and improving robustness. {Many of these recent methods \cite{lin2024clipose,chen2024secondpose,lin2024instance} require RGB images as input. However, in real-world applications, especially in challenging scenarios such as industrial environments \cite{dong2019ppr}, objects may lack distinctive color or texture, and lighting conditions can be highly variable. In contrast, the geometric information provided by depth images is generally more robust to changes in lighting and texture \cite{li2023depth}. Consequently, there are recent works \cite{chen2021fs,di2022gpv,zheng2023hs} focus on depth-only pose estimation to improve reliability in challenging conditions. Our method also follows this depth-only paradigm.}

Despite these advancements, challenges remain. The aforementioned methods only treat poses as training objectives, resulting in fragmented point cloud features that fail to explicitly capture the continuous nature of 6D poses (examples in Fig.\ \ref{p_3}), leading to inconsistencies in predictions and reduced generalization to unseen objects. While no existing works have explicitly addressed this issue for category-level pose estimation, several efforts have been made. \cite{wohlhart2015learning,balntas2017pose,zakharov20173d} 
adapted triplet loss \cite{weinberger2009distance} to learn rotation-aware representations from RGB-D images for 3D object pose estimation. However, these approaches are not directly applicable to category-level pose estimation, as they focus solely on rotations while neglecting translations, which are essential for predicting full 6D poses. Moreover, existing studies \cite{khosla2020supervised, chen2020simple} have demonstrated that triplet loss generally underperforms contrastive loss in representation learning tasks.
Recently, \cite{zha2024rank} proposed a ranking-based contrastive loss to learn representations for regression tasks while preserving the intrinsic continuity of task-specific targets. This method first ranks samples according to their targets and then contrasts them based on their relative rankings. Although it has shown superior performance, it is inherently limited to handling data with a single task type and does not incorporate category information. Consequently, it struggles to extend to category-level pose estimation, which involves multi-tasks (both rotation and translation) and multi-category data.

To capture the intrinsic continuity of 6D poses for category-level pose estimation and overcome the limitations of prior works \cite{wohlhart2015learning,balntas2017pose,zakharov20173d,zha2024rank}, we propose HRC-Pose, a novel {depth-based} framework that introduces a hierarchical ranking contrastive learning module that can learn multi-category point cloud representations while preserving both rotational and translational continuity. Our contrast learning strategy is based on a hierarchical ranking scheme. As illustrated in Fig.\ \ref{p_1}, we establish negative pairs for a randomly selected positive pair within a batch by first jointly ranking point clouds based on both rotational and translational differences. To ensure consistency in this joint ranking, we exclude point clouds where the ranking order defined by rotational differences conflicts with that defined by translational differences when compared to an anchor. After constructing negative pairs from this joint ranking, we then separately rank all point clouds based on either rotational or translational differences to introduce additional negative pairs. The resulting rotational and translational negative pairs are subsequently combined with the joint negative pairs to contrast point clouds and facilitate the learning of representations that preserve continuity in both rotation and translation, as shown in Fig.\ \ref{p_3}. Additionally, to handle multi-category scenarios, we contrast point clouds within each category independently and aggregate the contrastive losses across all categories. The learned rotational and translational representations are separately used for final category-level pose estimation.



The main contributions of this work are as follows: 
\begin{itemize}
\item We propose hierarchical ranking contrastive learning, {which is based on our} 6D pose-aware hierarchical ranking scheme to extract multi-category point cloud features {while preserving the continuity of both rotation and translation in the learned feature space.}
\item {We develop HRC-Pose, a depth-based framework for category-level object pose estimation that decouples pose into rotation and translation components, which are separately encoded, learned, and utilized through dedicated contrastive learning and pose estimation modules.}
\item {Extensive experiments demonstrate that HRC-Pose achieves state-of-the-art performance among depth-based methods} while running in real-time.
\end{itemize}

\section{RELATED WORKS}
\label{sec:related}

\begin{figure*}[h!] 
\begin{center}
\includegraphics[width=\textwidth]{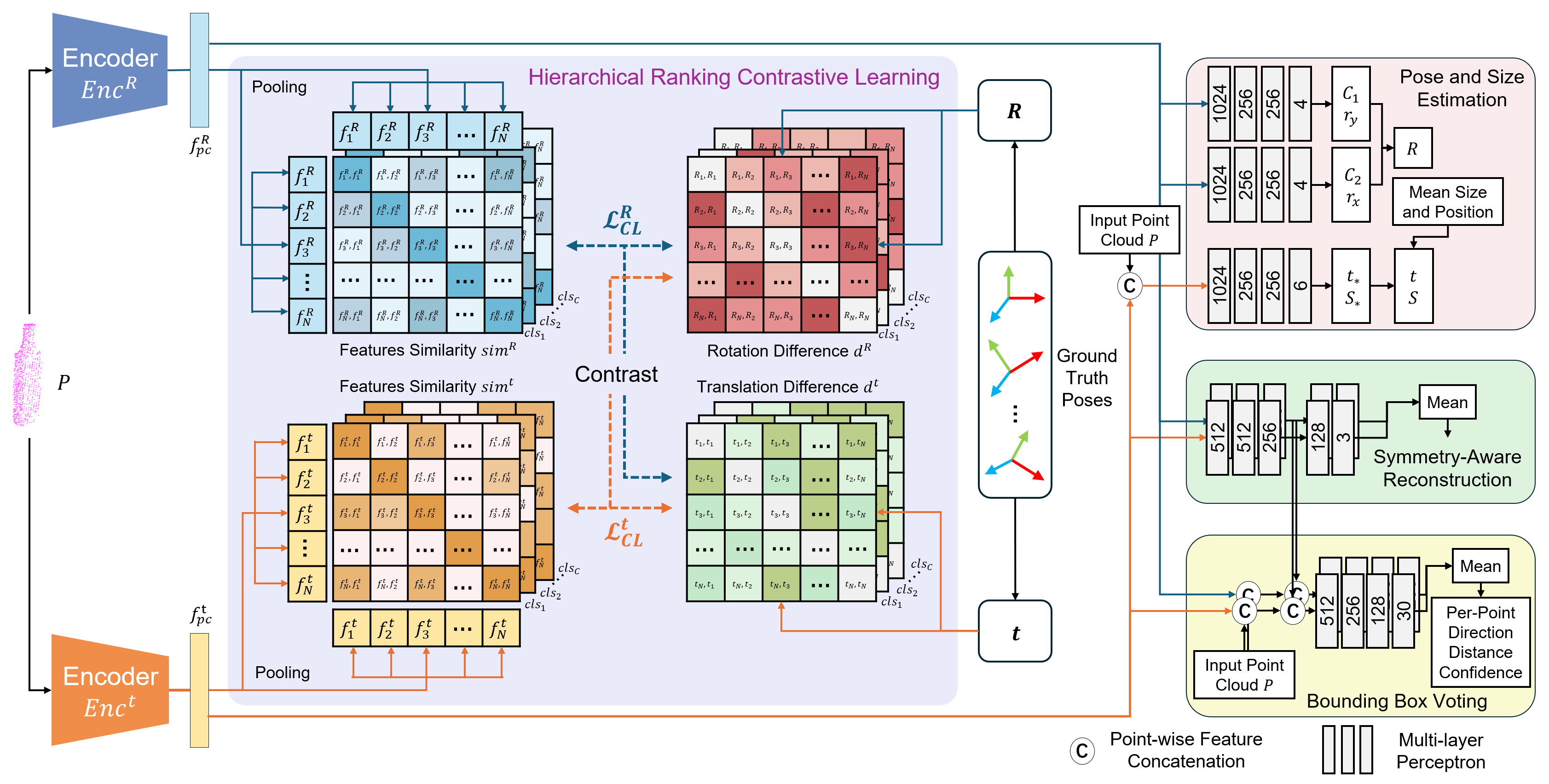}
\caption{{Overview of HRC-Pose.} 
Given point clouds $P$ and category labels $cls$, point cloud encoders $Enc^R$ and $Enc^t$ are trained to extract pose-aware embeddings w.r.t. ground truth poses $\boldsymbol{R}$ and $\boldsymbol{t}$ via our hierarchical ranking contrastive learning module.
Two contrastive learning objectives, $\mathcal{L}^{R}_{{CL}}$ and $\mathcal{L}^{t}_{{CL}}$, are introduced to contrast point cloud embeddings and 6D poses, enabling the learned representations $f^R_{pc}$ and $f^t_{pc}$ to capture the continuity underlying $\boldsymbol{R}$ and $\boldsymbol{t}$. 
Finally, $f^R_{pc}$ and $f^t_{pc}$ are fed into sub-modules for pose regression, symmetry-aware point cloud reconstruction, and bounding box voting.}
\label{p_2}
\end{center}
\vspace{-0.8cm}
\end{figure*}

\paragraph{Category-Level Object Pose Estimation}
Category-level object pose estimation predicts poses for unseen objects from predefined categories. Significant intra-class variations in shape and texture make this a challenging problem. A common approach involves building correspondences between observed objects and their canonical shapes, followed by pose estimation using the Umeyama algorithm \cite{umeyama1991least}. To build a canonical coordinate space invariant to pose and scale variations, \cite{wang2019normalized} proposes a novel normalized object coordinate space (NOCS) for each object category by mapping the 3D object model onto RGB images. Subsequent works \cite{wang2021category, feng2020fully, lee2021category, wang2022attention, corsetti2023revisiting, chen2023stereopose} have focused on fully leveraging the inherent relationships between RGB images and point clouds to predict the observed objects in their NOCS. In addition, other methods \cite{tian2020shape, chen2021sgpa, li2023generative} leverage prior category knowledge, such as geometric information, to aid pose estimation. Alternatively, some methods directly regress poses from observed point clouds without building correspondences to canonical shapes. For instance, \cite{chen2020category} adopts an analysis-by-synthesis framework that iteratively refines poses, while others \cite{chen2021fs, lin2021dualposenet, di2022gpv, zheng2023hs} have introduced advanced pose decoders to boost accuracy. Recently, generative models \cite{ho2020denoising, song2021maximum} have also been explored for this task \cite{diffusionnocs}. However, these models come with efficiency trade-offs. In contrast, our work aims to retain the efficiency of regression-based models while improving pose estimation accuracy. 

\paragraph{Contrastive Learning}
Contrastive learning \cite{le2020contrastive} has played an important role in representation learning, aiming to bring similar data points closer in feature space while pushing dissimilar ones apart. By comparing input data across the same or different modalities, contrastive learning facilitates the learning of robust feature representations. Many existing methods have introduced contrastive loss functions well-suited for self-supervised classification tasks \cite{dyer2014notes,oord2018representation,chen2020simple,wang2021contrastive}, while others \cite{khosla2020supervised,zheng2021weakly} extend these approaches to the supervised setting, further enhancing feature discriminability by effectively incorporating label information. Contrastive learning has also been explored in pose estimation-related tasks. \cite{wohlhart2015learning,balntas2017pose,zakharov20173d,xiao2021posecontrast} leverage contrastive learning to learn rotation-aware representations for 3D object pose estimation by contracting objects according to rotational differences. However, these works are not directly applicable to 6D pose estimation, which also requires modeling translations. More recently, \cite{lin2024clipose} applies contrastive learning to category-level object pose estimation, leveraging external semantic knowledge from image and text modalities to develop better learning of object category information. In contrast, our proposed model employs contrastive learning to explicitly learn representations that capture the intrinsic continuity of 6D poses. A direct related work to our approach is Rank-N-Contrast (RNC) \cite{zha2024rank}, which introduces a contrastive loss for regression tasks by contrasting samples based on their rankings in the target space. The learned representations exhibit continuity in the embedding space according to the continuous targets. However, RNC is limited to handling a single continuous variable and does not incorporate category information, making it unsuitable for category-level object pose estimation, which requires learning continuity in both rotations and translations while preserving category distinctions. Compared to RnC, our approach handles multi-task, multi-category scenarios, making the learning task significantly more challenging.

\section{METHODOLOGY}
Given an RGB-D image, the goal of category-level object pose estimation is to predict the 6D object pose, comprising rotation $R \in SO(3)$ and translation $t \in \mathbb{R}^3$, and size $s \in \mathbb{R}^3$. Similar to existing depth-based baselines like \cite{di2022gpv,zheng2023hs}, objects are segmented from the RGB image using object segmentation network, such as Mask-RCNN \cite{he2017mask}. The segmented object's depth image is back-projected into a point cloud $P$, which is then uniformly sampled to a fixed number $N_p$. Given the observed target point cloud $P \in \mathbb{R}^{N_p \times 3}$, we follow \cite{chen2021fs} to augment the point cloud and represent $R$ using two plane normals, $\boldsymbol{r}_x$ and $\boldsymbol{r}_y$. During inference, our proposed depth-based method exclusively uses point cloud data as input, without relying on RGB data.

\subsection{Feature Extractor}
Given the input point cloud $P$, we employ separate point cloud encoders, denoted as $Enc^R$ and $Enc^t$, to extract representations associated with rotation and translation. As shown in Fig.\ \ref{p_2}, these encoders produce point-wise embeddings $f^R_{pc}\in\mathbb{R}^{{N_p}\times{D}}$ and $f^t_{pc}\in\mathbb{R}^{{N_p}\times{D}}$, respectively. For contrastive learning, pooling layers are applied after $Enc^R$ and $Enc^t$ to aggregate global point cloud embeddings, denoted as $f_i^R$ and $f_i^t$ from $f^R_{pc}$ and $f^t_{pc}$, respectively.

\subsection{Hierarchical Ranking Contrastive Learning}
With the global point cloud embeddings extracted from the encoders, we utilize them in our proposed hierarchical ranking contrastive learning. 
This contrastive learning strategy captures the intrinsic continuity underlying multiple task-specific targets ($R$ and $t$ in our experiments) while incorporating multi-category information through the 6D pose-aware hierarchical ranking introduced below. 


\paragraph{6D Pose-Aware Hierarchical Ranking.} Given a randomly sampled batch of $N$ point clouds from the same category and their corresponding 6D poses, represented as $\{(p_n,y^g_n)\}_{n\in[N]}$,$g \in \{R,t\}$, where $p$ denotes a point cloud, $y^g$ denotes its rotation or translation values, we select an anchor point cloud indexed by $i$ with its feature $f_i^g$. For simplicity, we use the index to refer to the corresponding point cloud within the batch. Positive pairs are defined as $(i, j)^{N}_{j=1,j\neq i}$. For a randomly selected positive pair $(i,j)$, its corresponding negative pairs $(i,k)_{k\in N, k\neq i}$ are those where $k$ has a higher rank in ascending order than $j$ w.r.t. the anchor $i$. To incorporate both rotation and translation, we propose a hierarchical ranking scheme to define corresponding negative pairs at two levels based on a given positive pair $(i,j)$, as illustrated in Fig.\ \ref{p_1}.

At the first level (red box in Fig.\ \ref{p_1}), we define \textbf{joint negative pairs} as $(i,k)_{k\in N, k\neq i}$ if both ${\Delta R}_{ij}< {\Delta R}_{ik}$ and ${\Delta t}_{ij}< {\Delta t}_{ik} $ hold simultaneously, where $\Delta$ is the distance measure between two poses ($R$ or $t$). In other words, these negative samples $k$ are ranked higher than $j$ w.r.t. the anchor $i$. These samples also play an important role in contrastive learning as \textit{strong negatives}, as they are selected under a strict criterion that captures the inherent continuity of both $R$ and $t$. We further define the set of negative sample features in this joint ranking {$S_{i, j}^{joint, g}:=\left\{f_k^g\right\}$}, where each {$f_k^g$} represents the {rotation-aware or translation-aware} feature of a corresponding strong negative sample $k$. 

At the second level (blue and orange boxes in Fig.\ \ref{p_1}), we define \textbf{rotational negative pairs} $(i,k_R)_{k_R\in N, k_R\neq i}$ and \textbf{translational negative pairs} $(i,k_t)_{k_t\in N, k_t\neq i}$ by separately considering ${\Delta R}$ and ${\Delta t}$. Given a positive pair $(i,j)$, if ${\Delta R}_{ij}< {\Delta R}_{ik_R}$ or ${\Delta t}_{ij}< {\Delta t}_{ik_t} $, then the samples $k_g$ are considered as negative w.r.t. $g \in \{R,t\}$. The corresponding feature sets are then defined as {$S_{i, j}^{g}:=\left\{f_{k}^{g}\right\}$}. Compared to $S_{i, j}^{joint, g}$, the set $S_{i, j}^{g}$ includes more features within the batch due to a more relaxed ranking criterion, thereby enhancing data utilization efficiency in contrastive learning. 

Incorporating both joint negative pairs and rotational/translational negative pairs ensures a balance between strict 6D pose continuity enforcement and efficient data utilization in contrastive learning. The joint set {$S_{i, j}^{joint, g}, g \in \{R,t\}$} consists of strong negatives, selected under the strictest ranking criterion where both rotation and translation differences must satisfy the ranking condition simultaneously. This allows the model to pay special attention to the most reliable negative samples. However, while joint negative pairs are highly reliable, they can be too restrictive, limiting the number of available negatives in a batch. To improve data efficiency, the rotational/translational-specific sets $S_{i, j}^{g}, g \in \{R,t\}$ expand the set of available negatives by relaxing the ranking constraints, allowing either rotation or translation to be considered independently. This strategy enhances contrastive learning by making full use of batch data while preserving the continuity of 6D poses.



\paragraph{Our Contrastive Learning Loss. }

With the negative pairs and feature sets {$S_{i, j}^{joint, g}$ and $S_{i, j}^{g}, g \in \{R,t\}$} defined in the aforementioned hierarchical ranking scheme, we first define {the joint contrastive loss function $l^{joint, g}$ based on $S_{i, j}^{joint, g}, g \in \{R,t\}$} as follows: 
\begin{equation}\label{eq1}
\begin{aligned}
\resizebox{.44 \textwidth}{!}{
$l_{(i)}^{joint, g}=\frac{1}{N-1} \sum_{j=1, j \neq i}^{N}-\log \frac{\exp \left(\operatorname{sim}\left(f_{i}^{g}, f_{j}^{g}\right) / \tau\right)}{\sum_{f_{k}^{g} \in \mathcal{S}_{i, j}^{joint, g}} \exp \left(\operatorname{sim}\left(f_{i}^{g}, f_{k}^{g}\right) / \tau\right)},$
}
\end{aligned}
\end{equation}
\begin{equation}\label{eq2}
\begin{aligned}
\resizebox{0.18\textwidth}{!}{
$
l^{joint, g} = \frac{1}{N} \sum_{i=1}^{N}l_{(i)}^{joint, g}, $
}
\end{aligned}
\end{equation}
where $\operatorname{sim}(\cdot)$ denotes the feature similarity measured by the negative $L_2$ norm, and $\tau$ is the temperature. 

Then we define rotational/translational contrastive loss function $l^{g}$ based on $S_{i, j}^{g}, g \in \{R,t\}$ as follows: 
\begin{equation}\label{eq1}
\begin{aligned}
\resizebox{.42 \textwidth}{!}{
$l_{(i)}^g=\frac{1}{N-1} \sum_{j=1, j \neq i}^{N}-\log \frac{\exp \left(\operatorname{sim}\left(f_i^g, f_j^g\right) / \tau\right)}{\sum_{f_k^g \in \mathcal{S}_{i, j}^g} \exp \left(\operatorname{sim}\left(f_i^g, f_k^g\right) / \tau\right)},$
}
\end{aligned}
\end{equation}
\begin{equation}\label{eq2}
\begin{aligned}
\resizebox{0.16\textwidth}{!}{
$
l^g = \frac{1}{N} \sum_{i=1}^{N}l_{(i)}^g$.
}
\end{aligned}
\end{equation}

Each aforementioned contrastive loss enforces the point cloud feature similarity between $i$ and $j$ to be larger than that between $i$ and any $k$ covered by the corresponding negative sample set $S_{i, j}$. By minimizing these loss functions, the feature similarity between $\boldsymbol{f}_i$ and the feature of the \textit{closest} sample in task space becomes the \textit{largest} within a batch, followed by the second closest, and so on. This enables the learned point cloud embeddings to capture the intrinsic order of their 6D poses.

Since our proposed hierarchical ranking scheme is applied to point clouds within the same category, the $l^{{joint}}$, $l^{{R}}$, and $l^{{t}}$ are defined per category $c$. To enable contrastive learning across multiple categories, we average each loss function over all categories during training: 
\begin{equation}\label{eq8}
\resizebox{0.40\textwidth}{!}{
$
\mathcal{L}^{{joint, g}}=\frac{1}{C} \sum_{c=1}^{C} l^{{joint, g}}_c, \ \ \ 
\mathcal{L}^{{g}}=\frac{1}{C} \sum_{c=1}^{C} l^{{g}}_c,$
}
\end{equation}
where $C$ is the number of categories. 

The final contrastive learning loss functions for $R$ and $t$ are defined as follows: 
\begin{equation}\label{eq3}
\begin{aligned}
\mathcal{L}^{R}_{{CL}} &= \mathcal{L}^{{joint, R}} + \lambda \mathcal{L}^{R}, \\
\mathcal{L}^{t}_{{CL}} &= \mathcal{L}^{{joint, t}} + \lambda \mathcal{L}^{t},
\end{aligned}
\end{equation}

where $\lambda$ is a weight to balance the loss terms. 

\paragraph{Connection to Rank-N-Contrast.}
RNC loss \cite{zha2024rank} can be seen as a special case of our hierarchical ranking contrastive loss when only one task and one category are considered. To overcome the limitations of RNC, which is designed for regression tasks with \textit{a single task and no category} information, we introduce a hierarchical ranking scheme that handles \textit{multi-task, multi-category} scenarios. Our scheme involves both strict continuity enforcement in 6D poses and efficient data utilization in contrastive learning. Additionally, we incorporate weighting factors into the loss terms to regulate their contributions to the final loss, ensuring a proper balance of negative pairs from different sets during training. Since our data spans multiple categories, we adopt a simple yet effective strategy that enforces contrast only among samples within the same category.

\subsection{Pose Estimation Modules {with Rotation-Aware and Translation-Aware Embeddings}}
{While the global point cloud embeddings are used in our contrastive learning to capture pose continuity, we further utilize the corresponding point-wise embeddings for category-level object pose estimation. Specifically, we propose to separately leverage the rotation-aware point-wise embeddings $f^R_{pc}$ and translation-aware point-wise embeddings $f^t_{pc}$ from $Enc^R$ and $Enc^t$, respectively. These features are independently fed into three sub-modules for pose regression, symmetry-aware point cloud reconstruction, and bounding box voting. Although prior works \cite{di2022gpv,zheng2023hs} incorporate similar modules in their architectures, they only use shared point-wise embeddings across all modules. In contrast, we argue that rotation and translation represent distinct aspects of object pose, and that treating them separately is beneficial.}

{As illustrated in Fig.\ \ref{p_2}, we explicitly predict the object's rotation $R$ using rotation-aware embeddings $f^R_{pc}$, which preserve rotation continuity, and predict translations $t$ using translation-aware embeddings $f^t_{pc}$, which preserve translation continuity. For the symmetry-aware reconstruction sub-module and the bounding box voting sub-modules, we process $f^R_{pc}$ and $f^t_{pc}$ in parallel and average their outputs to obtain the final results.}

By combining the losses in contrastive learning and pose estimation modules, the overall loss function is as follows:
\begin{equation}\label{eq3}
\begin{aligned}
\resizebox{0.48\textwidth}{!}{
$
\mathcal{L}_{overall} = \mathcal{L}^{R}_{CL} + \mathcal{L}^{t}_{CL} + \lambda_{B a s i c} \mathcal{L}^{B a s i c}+\lambda_{B B} \mathcal{L}_{(R, t, s)}^{B B}+\lambda_{P C} \mathcal{L}_{(R, t, s)}^{P C},
$
}
\end{aligned}
\end{equation}
where $\mathcal{L}^{R}_{CL}$ and $\mathcal{L}^{t}_{CL}$ are as described above. $\mathcal{L}^{B a s i c}$, $\mathcal{L}_{(R, t, s)}^{B B}$ and $\mathcal{L}_{(R, t, s)}^{P C}$ are losses defined in \cite{di2022gpv} related to the pose estimation modules.
\section{EXPERIMENTS}

\subsection{Experimental Setup}
\paragraph{Datasets.}
We conducted our experiments using the widely adopted NOCS dataset \cite{wang2019normalized}, which includes six object categories: \textit{bottle}, \textit{bowl}, \textit{camera}, \textit{can}, \textit{laptop}, and \textit{mug}. We followed \cite{tian2020shape} to process the dataset and split it into training and testing sets. The dataset is generally divided into two subsets: CAMERA and REAL. The CAMERA subset consists of 275K synthetic rendered images for training and 25K for testing, generated using 1,085 object models from ShapeNetCore \cite{chang2015shapenet}. The testing set from the CAMERA subset is referred to as CAMERA25. The REAL subset is more challenging, containing real-world scene images of the same six categories. It includes three unique instances per category for both the training and testing sets. Specifically, it consists of seven training scenes with 4.3K images and six testing scenes with 2.75K images. The REAL testing set is referred to as REAL275.

\paragraph{Metrics.}
Following \cite{zhang2022rbp,di2022gpv, zheng2023hs}, we compute the mean Average Precision (mAP) in terms of $n^{\circ}$ and $m$ cm to evaluate pose estimation accuracy, where $n$ and $m$ represent the prediction errors for rotation and translation, respectively. Specifically, the predicted rotation error is less than $n^{\circ}$ and the predicted translation error is less than $m$ cm. We use evaluation metrics of $5^{\circ}2cm$, $5^{\circ}5cm$, $10^{\circ}2cm$ and $10^{\circ}5cm$. Following \cite{wang2019normalized}, we ignore the rotation error around the symmetry axis for symmetric objects such as bottles, bowls, and cans. For the mug category, it is treated as a symmetric object when the handle is not visible and as a non-symmetric object when the handle is visible. We also report the mean precision of 3D intersection over union (IoU) at various thresholds of 50\% and 75\% to evaluate the size prediction.

\paragraph{Implementation Details.}
For a fair comparison, we employ the same data preprocessing as previous depth-based works \cite{di2022gpv, zheng2023hs} to achieve the point cloud data with 1,024 points as the direct input of our network. Data augmentation strategies are adapted from \cite{chen2021fs}, including random scaling, uniform noise, rotational and translational perturbations, and adjustments based on 3D bounding boxes.

In 6D pose-aware hierarchical ranking, for the rotation distance measure $\Delta R$, we follow \cite{chen2021fs} and decouple the rotation matrix $\boldsymbol{R} \in SO(3)$ into two perpendicular plane normals $\boldsymbol{r}_x$ and $\boldsymbol{r}_y$. We then define 
\begin{equation}
\resizebox{.433 \textwidth}{!}{
$
\Delta R_{ij}= (1 - cos_{sim}(\boldsymbol{r}_{xi}, \boldsymbol{r}_{xj})) + (1 - cos_{sim}(\boldsymbol{r}_{yi}, \boldsymbol{r}_{yj}))$,
}
\end{equation}
where $cos_{sim}(\cdot, \cdot)$ denotes the cosine similarity. Moreover, we ignore the rotation distance around the symmetry axis for symmetric objects, {following \cite{chen2021fs}.}

For the translation distance measure $\Delta t$, we define it as 
\begin{equation}
\resizebox{.15 \textwidth}{!}{
$
\Delta t_{ij} = MSE(\boldsymbol{t}_i, \boldsymbol{t}_j)$},
\end{equation}
where $MSE$ represents the mean squared error that computes the average of the squared differences between corresponding elements of two vectors.

\begin{figure}[t]
\center
\includegraphics[width=0.46\textwidth]{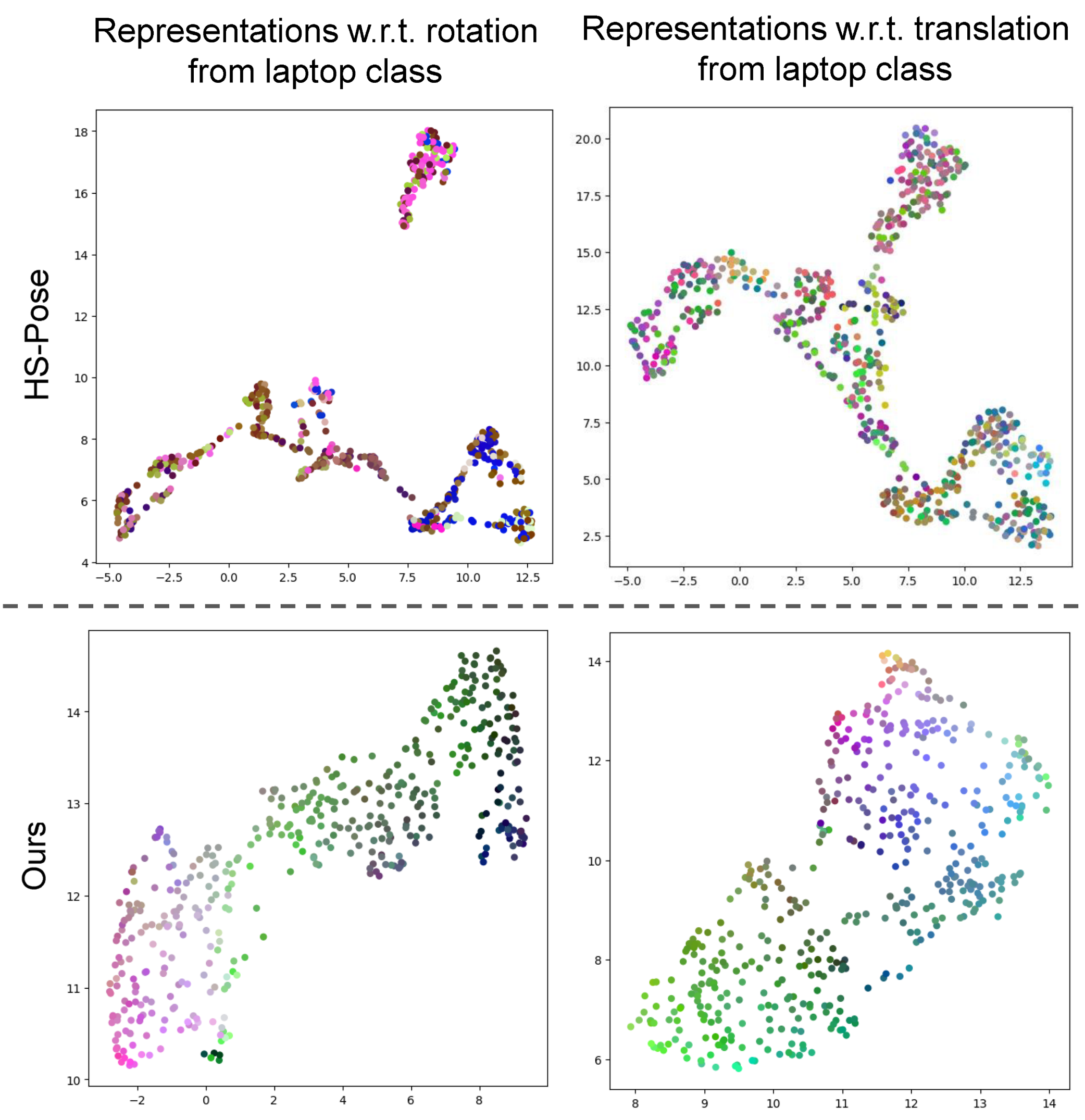}
\caption{{UMAP visualizations of laptop representations w.r.t. rotation and translation on the REAL275 dataset.} We visualize the representations for pose estimation from HS-Pose and from the trained encoders of our contrast learning module via UMAP. The RGB color of each point is determined by the corresponding rotation and translation values.}
\label{p_3}
\end{figure}

\begin{figure}[t]
\includegraphics[width=0.48\textwidth]{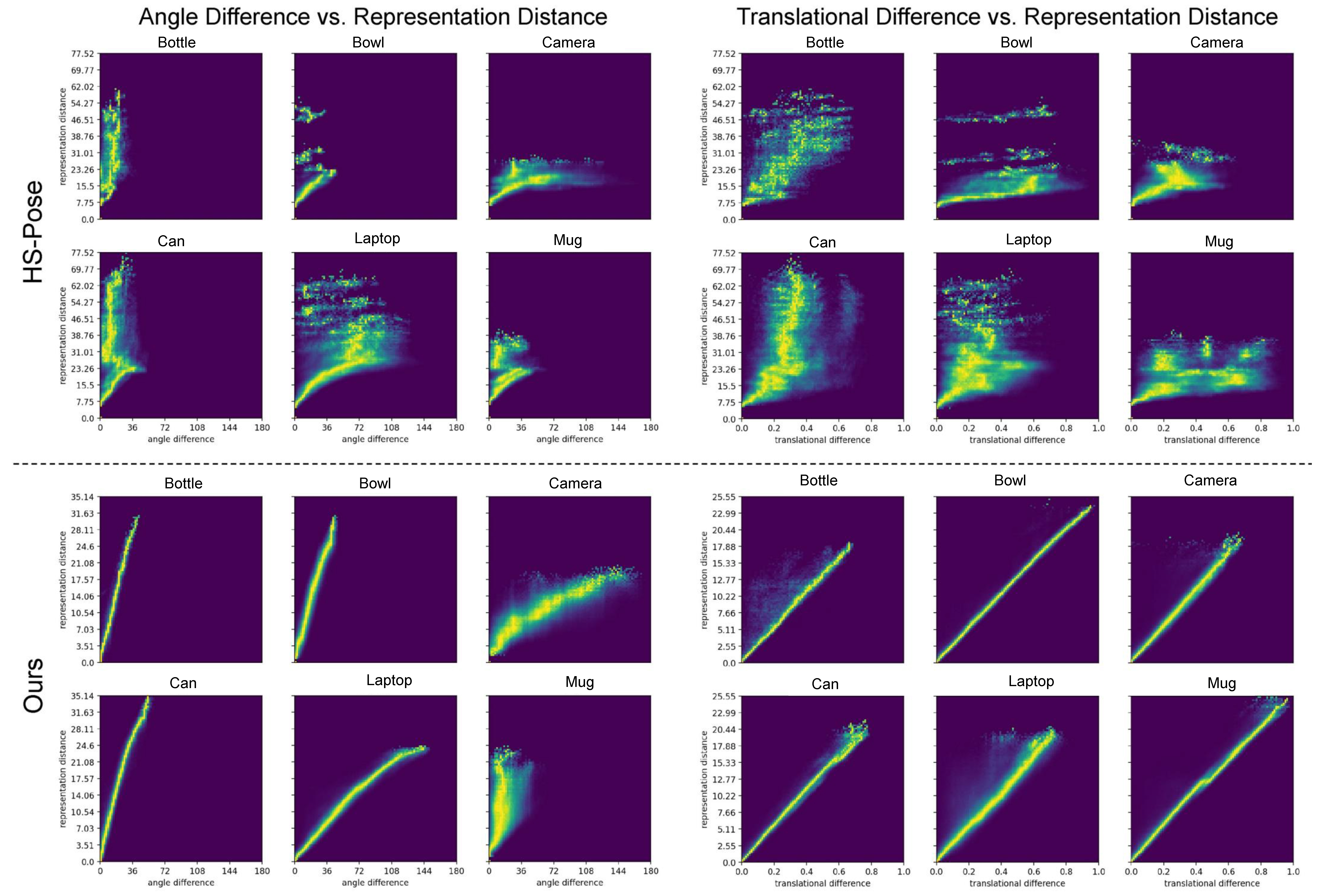}
\caption{Illustration of the correlations between pose differences and representation distances for HS-Pose and our method on the REAL275 dataset.}
\label{p_4}
\end{figure}

\begin{figure}[t]
\includegraphics[width=0.46\textwidth]{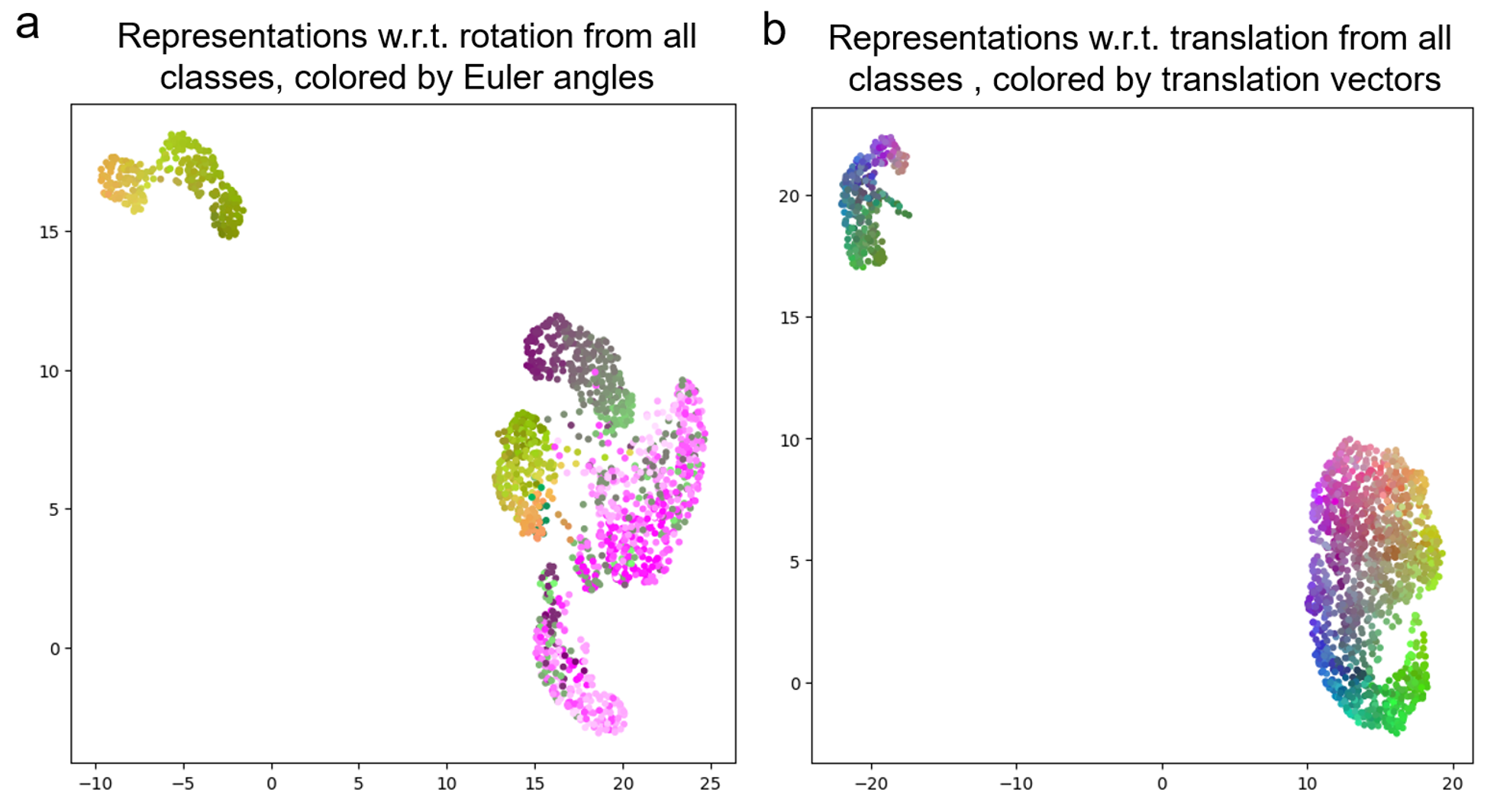}
\caption{UMAP visualizations of point cloud representations from our $Enc^R$ and $Enc^t$ on the REAL275 dataset. In \textbf{a} and \textbf{b}, the RGB color of each point is determined by scaling the corresponding rotation and translation values, respectively.}
\label{p_5}
\end{figure}

For our hierarchical ranking contrastive learning loss, we set $\tau$ to 2, following \cite{zha2024rank}. We adapt AdamW as the optimizer, with a weight decay of 0.001 and the ReduceLROnPlateau learning rate scheduler (patience = 10, factor = 0.5). For feature extraction, we employ 3D-GCN with HS layers \cite{zheng2023hs} as the backbone of $Enc^R$ and $Enc^t$ to obtain point-wise features $f_{pc}^R$ and $f_{pc}^t$, where each point's feature dimension is 512. For contrastive learning, $f_{pc}^t$ is concatenated with each point's coordinate. We adopt maxpooling layers to obtain global features $f^R$ and $f^t$. The weighting factor $\lambda$ is 0.8. The training strategy and hyperparameters for our three sub-modules follow \cite{di2022gpv,zheng2023hs}. All experiments are conducted on a single Nvidia L40s GPU, and all categories are trained together with a batch size of 32. During inference, we follow the standard evaluation protocol by setting the batch size to 1 for FPS computation and excluding data preprocessing time, consistent with prior works \cite{di2022gpv,zheng2023hs}.

\begin{table*}
\small
\centering
\renewcommand{\arraystretch}{1.0}
\resizebox{\textwidth}{!}{
\begin{tabular}{c|ccc|cc|cccc|cc|cccc}
\hline
\multirow{2}{*}{Method} & Data & \multirow{2}{*}{Prior} & External & \multicolumn{6}{c|}{REAL275} & \multicolumn{6}{c}{CAMERA275}  \\
\cline{5-16}
& Type & & Data & IoU$_{50}$ & IoU$_{75}$ & $5^{\circ} 2cm $ & $5^{\circ} 5cm$ & $10^{\circ} 2cm $ & $10^{\circ} 5cm$ &  IoU$_{50}$ & IoU$_{75}$ & $5^{\circ} 2cm$ & $5^{\circ} 5cm$ & $10^{\circ} 2cm$ & $10^{\circ} 5cm$  \\
\hline
SPD \cite{tian2020shape} & RGB-D & \cmark & \xmark &  77.3 & 53.0 & 19.3 & 21.4 & 43.2 & 54.1 & 93.2 & 83.1 &54.3 & 59.0 & 73.3 & 81.5  \\
DualPoseNet \cite{lin2021dualposenet} & RGB-D & \xmark & \xmark & 79.8 & 62.2 & 29.3 & 35.9 & 50.0 & 66.8 & 92.4 & 86.4 &64.7 & 70.7 & 77.2 & 84.7  \\
SGPA \cite{chen2021sgpa} & RGB-D & \cmark & \xmark & 80.1 & 61.9 & 35.9 & 39.6 & 61.3 & 70.7 &  93.2 & 88.1 &70.7 & 74.5 & 82.7 & 88.4  \\
DPDN \cite{lin2022category} & RGB-D & \cmark & \xmark &  83.4 & 76.0 & 46.0 & 50.7 & 70.4 & 78.4 & -& -& -& -& -& - \\ 
IST-Net \cite{liu2023net} & RGB-D & \xmark & \xmark &  82.5 & 76.6 & 47.8 & 55.1 & 69.5 & 79.6 &  93.7 & 90.8 &71.3 & 79.9 & 79.4 & 89.9  \\
CLIPose \cite{lin2024clipose} & RGB-D & \xmark & \cmark & - & - & 48.7 & 58.3 & 70.4 & 85.2 &  - & - & 74.8 & 82.2 & 82.0 & 91.2  \\
AG-Pose \cite{lin2024instance} & RGB-D & \xmark & \xmark & 83.7 & {79.5} & 54.7 & {61.7} & {74.7} & 83.1 &  93.8 & 91.3 &77.8 & 82.8 & 85.5 & 91.6  \\
SecondPose \cite{chen2024secondpose} & RGB-D & \cmark & \cmark & - & - & 56.2 & 63.6 & {74.7} & 86.0 & -& -& -& -& -& - \\
\hline
FS-Net \cite{chen2021fs} & D & \xmark & \xmark & 81.1 & 52.0 & 19.9 & 33.9 & - & 69.1 & -& -& -& -& -& - \\
GPV-Pose \cite{di2022gpv} & D & \xmark & \xmark &  \underline{83.0} & 64.4 & 32.0 & 42.9 & 55.0 & 73.3 & -& -& -& -& -& - \\
SAR-Net \cite{lin2022sar} & D & \cmark & \xmark & 79.3 & 62.4 & 31.6 & 42.3 & 50.4 & 68.3 & 86.8 & 79.0 &66.7 & 70.9 & 75.3 & 80.3  \\
SSP-Pose \cite{zhang2022ssp} & D & \cmark & \xmark &  82.3 & 66.3 & 34.7 & 44.6 & - & 77.8 &  - & 86.8 &64.7 & 75.5 & - & 87.4  \\
RBP-Pose \cite{zhang2022rbp} & D & \cmark & \xmark & - & 67.8 & 38.2 & 48.1 & 63.1 & 79.2 &  93.1 & 89.0 &\underline{73.5} & 79.6 & \underline{82.1} & \underline{89.5}  \\
HS-Pose \cite{zheng2023hs} & D & \xmark & \xmark &  82.1 & \underline{74.7} & \underline{46.5} & \underline{55.2}  & \underline{68.6} & \underline{82.7} & \underline{93.3} &\underline{89.4} &73.3 & \underline{80.5} & 80.4 & {89.4}  \\
\hline
Ours  & D & \xmark & \xmark & \textbf{83.4} & \textbf{77.8} & \textbf{49.8} & \textbf{58.6} & \textbf{72.5} & \textbf{85.4} &  \textbf{93.6} & \textbf{90.0} & \textbf{75.2} & \textbf{82.5} & \textbf{82.3} & \textbf{91.2}  \\ 
\hline
\end{tabular}
}
\caption{{Comparison with state-of-the-art methods on the REAL275 {and CAMERA25} dataset.} For depth-only methods, the best results are in bold, and the second-best results are underlined. \textit{Data Type} denotes the method's input data type {for both training and inference.} \textit{Prior} means whether the method requires category priors. {\textit{External Data} indicates whether the method requires external training data.} }
\label{t1}
\end{table*}

\subsection{Comparison with State-of-the-Art Methods}
\paragraph{Analysis on representations from our contrastive learning module.} To analyze the learned representations from our $Enc^R$ and $Enc^t$, we visualize them using UMAP \cite{mcinnes2018umap} in the second row of Fig.\ \ref{p_3}, where each point indicates the learned embeddings from a point cloud and the RGB color represents its pose. For comparison, we also visualize the learned representations for pose regression by the most competitive depth-based baseline, HS-Pose \cite{zheng2023hs}, in the first row of Fig.\ \ref{p_3}. All representations are trained on the REAL275 dataset. In the first column, the RGB color of each point from $Enc^R$ w.r.t. rotation is scaled based on the corresponding Euler angles (ignoring rotation around symmetry axes for symmetric objects). In the second column, the RGB color of each point from $Enc^t$ w.r.t. translation is scaled based on the translation $XYZ$ values. From the visualizations in Fig.\ \ref{p_3}, we observe that the representations learned by HS-Pose are fragmented, with colors exhibiting random-like patterns, indicating its inability to capture the continuous relationships underlying point cloud poses. In contrast, our learned representations are more continuous, with colors forming distinct rainbow-like patterns. 

Additionally, we analyze the correlations between pose (rotation or translation) differences and the Euclidean distances between learned representations by plotting correlation maps \cite{balntas2017pose} of features from HS-Pose and our $Enc^R$ and $Enc^t$, as shown in Fig.\ \ref{p_4}. The compact, diagonal-like shapes of our curves show that the distances of the representations learned from our method are significantly more correlated with their pose differences compared to HS-Pose. Regarding Pearson correlation coefficients (P), our method achieves an average P of 0.92, compared to 0.20 for HS-Pose. These results confirm that our method produces representations whose pairwise distances align more closely with 6D pose differences, validating the effectiveness of our contrastive learning strategy in capturing the intrinsic continuity and ordering in pose space.

Furthermore, we visualize point cloud representations across all categories learned by our contrastive learning module on the REAL275 dataset using UMAP, as shown in Fig.\ \ref{p_5}. Fig.\ \ref{p_5}a and Fig.\ \ref{p_5}b show the representations from $Enc^R$ and $Enc^t$, respectively, with point colors assigned in the same manner as in Fig.\ \ref{p_3}. Notably, the scales for computing RGB colors in Fig.\ \ref{p_5} are based on poses across all categories, whereas the scales in Fig.\ \ref{p_3} are based on the poses within the laptop class. Consequently, the RGB colors do not directly match between the two figures for the same class. From Fig.\ \ref{p_5}, we observe that our learned point cloud representations exhibit clear continuity with respect to both rotation and translation, even across multiple object categories. This observation demonstrates that our contrastive learning strategy effectively captures continuous pose-aware representations that preserve the intrinsic 6D pose relationships among multi-category point clouds.

\paragraph{Performance on REAL275 dataset.}
In Table \ref{t1}, which reports mAP scores across various metrics, we compare with state-of-the-art methods on the REAL275 dataset. As REAL275 is collected in real-world scenes, it is more susceptible to various types of noise, making it a better benchmark for evaluating the robustness of pose estimation methods against real-world interference.
Without using any shape prior information, our method outperforms all depth-based methods across all metrics and even surpasses most RGB-D-based baselines. Moreover, our method outperforms CLIPose \cite{lin2024clipose}, which leverages contractive learning with external semantic knowledge from image and text modalities. In contrast, we only use the data in REAL275, yet achieve better performance by using contractive learning to capture the intrinsic continuity of 6D poses. 
It is worth noting that our proposed contrastive learning strategy introduces no additional cost during inference, allowing our model to maintain efficiency when adopting an efficient point cloud encoder. In terms of inference speed, our method achieves 122.6 FPS, comparable to the second-best depth-based method, HS-Pose (121.5 FPS), when tested on the same machine, thus enabling real-time execution. These results highlight the effectiveness of our contrastive learning strategy and overall framework design. Additionally, our model’s robustness to noise comes from its point cloud encoder, which mitigates the influence of outliers \cite{zheng2023hs}.

\paragraph{Performance on CAMERA25 dataset.}
Table \ref{t1} also reports the performance comparison on the CAMERA25 dataset. Unlike REAL275, CAMERA25 is collected in simulated environments and contains no noise, making it less practical and less challenging. As shown in the results, all baseline methods perform significantly better on CAMERA25 than on REAL275, and the performance of recent baselines appears to have saturated on this benchmark. Despite this, our method achieves the best results among depth-based methods and even outperforms some RGB-D-based methods, further demonstrating the effectiveness of our contrastive learning module and overall framework. 

\begin{table}
\resizebox{0.47\textwidth}{!}{
\begin{tabular}{c|c|c c c c}
\hline
Group & Method & $5^{\circ} 2cm$ & $5^{\circ} 5cm$ & $10^{\circ} 2cm$ & $10^{\circ} 5cm$  \\
\hline
\multirow{3}{*}{1} & w/o {CL} & 46.5 & 55.2 & 68.6 & 82.7 \\
 & w/o translation in {CL} & 47.2 & 57.0 & 70.8 & 83.7 \\
  & w/o rotation in {CL} & 48.1 & 58.1 & 71.2 & 84.1 \\
\hline
\multirow{2}{*}{2} & Only $\mathcal{L}^{\text{joint}}$ & 47.4 & 56.0 & 69.5 & 83.4 \\
 & W/O $\mathcal{L}^{\text{joint}}$ & 48.5 & 57.6 & 72.5 & 84.3 \\
\hline
\multirow{2}{*}{3} & w/o our categorical design & 43.3 & 51.2 & 64.1 & 76.2 \\
 & Use SupCon loss \cite{khosla2020supervised} & 42.8 & 52.0 & 63.4 & 76.0 \\
\hline
\multirow{4}{*}{4} & Use 3D-GCN w {CL} & 36.9 & 47.0 & 61.2 & 75.6 \\
& Use PointNet++ w {CL} & 32.1 & 42.6 & 56.1 & 74.2 \\
& Use 3D-GCN w/o {CL} & 30.6 & 39.1 & 50.7 & 62.7 \\
& Use PointNet++ w/o {CL} & 25.6 & 34.1 & 44.8 & 59.3 \\
\hline
5 & Feed global features into sub-modules & 40.3 & 50.1 & 62.9 & 78.2 \\
\hline
\multirow{2}{*}{6} & Direct concatenation & 47.0 & 55.3 & 68.3 & 83.0  \\
 & Use cross attention Layers & 46.8 & 55.6 & 68.5 & 82.5\\
\hline
7 & Ours & \textbf{49.3} & \textbf{59.1} & \textbf{74.0} & \textbf{86.8} \\
\hline
\end{tabular}
}
\caption{{Ablation studies on the REAL275 dataset.} {CL} denotes our contrastive learning module.}
\label{t3}
\end{table}

\subsection{Ablation Studies}
To verify the effectiveness of our proposed method, we conducted ablation studies on the REAL275 dataset. Specifically, we evaluated the impact of our contrastive learning module, the designs of the contrastive learning losses, the choice of point cloud encoder, and the characteristics of the point cloud features. The complete ablation results are presented in Table \ref{t3}.

\textbf{Effect of the contrastive learning module.\ }
We verify the efficacy of our hierarchical contrastive learning module in Groups 1 and 4 of Table \ref{t3}. Removing the entire module or any part of it from our model, as well as from its 3D-GCN and PointNet++ variants, leads to notable performance drops. These results demonstrate that our contrastive learning strategy is effective regardless of encoder strength and highlight its effectiveness in guiding the point cloud encoder for more accurate pose estimation.

\textbf{\noindent Effect of the hierarchical design in contrastive loss.\ }
We examine the impact of the hierarchical design in our contrastive learning loss in Group 2 of Table \ref{t3}. Using only the joint loss $\mathcal{L}^{{joint}}$ in contrastive learning significantly degrades performance, likely because the proportion of strong negative pairs within a batch is relatively small. Additionally, we remove $\mathcal{L}^{{joint}}$ and train only with $\mathcal{L}^{{g}}, g \in \{R,t\}$. This setup does not perform as well as our full model, suggesting that the joint loss helps enforce pose-aware continuity in the learned representations.  

\textbf{Effect of the categorical design in contrastive loss.\ }
We evaluate the categorical design in Group 3 of Table \ref{t3}. Removing the categorical information and applying a single loss across all categories results in a significant performance drop. Additionally, we replace our categorical design with the SupCon loss from \cite{khosla2020supervised}, which is designed for classification tasks, and sum it with the single loss term. However, this setup also underperforms compared to averaging single-category losses across all categories, indicating the effectiveness of our categorical design.

\textbf{Effect of the point cloud encoder type.\ }
We analyze the impact of different point cloud encoders in Group 4 of Table \ref{t3}. Replacing the 3D-GCN with HS layers by a regular 3D-GCN leads to reduced performance. Substituting it with PointNet++ \cite{qi2017pointnet++} results in a significant performance drop, highlighting the effectiveness of our chosen encoder.

\textbf{Effect of the point cloud feature type.\ }
We test the effect of feeding global point cloud features into sub-modules in Group 5 of Table \ref{t3}. The performance drops considerably, likely due to global features lacking sufficient spatial details required for accurate pose estimation.

\textbf{Effect of point cloud feature processing in sub-modules.\ }
In Group 6 of Table \ref{t3}, we examine different point cloud feature processing methods within the sub-modules. Instead of processing them separately, we first concatenate the features and then feed them into the reconstruction and voting sub-modules. We also explore using an additional cross-attention layer for feature fusion. However, both approaches result in suboptimal performance. This confirms that our strategy of separately processing the features and averaging the outputs yields the best results.
\section{CONCLUSION}
\label{sec:con}
In this work, we present a novel depth-based framework for category-level object pose estimation. Our approach introduces a hierarchical ranking contrastive learning module to extract multi-category point cloud representations that can capture the intrinsic continuity of 6D poses. The learned representations enhance pose estimation accuracy, achieving state-of-the-art performance among depth-based methods on the REAL275 and CAMERA25 datasets while maintaining real-time inference speed. This makes our method well-suited for real-world applications. For future work, our framework can be expanded to process RGB-D images, providing opportunities for further research in this area. Additionally, our hierarchical ranking contrastive learning framework is general enough to be applied to other regression tasks involving multiple targets.
\newpage
{
    \small
    \bibliographystyle{ieeenat_fullname}
    \bibliography{main}
}

\end{document}